# Development and evaluation of an Explainable Prediction Model for Chronic Kidney Disease Patients based on Ensemble Trees

Pedro A. Moreno-Sanchez [1][0000-0003-3987-1789]

[1] School of Health Care and Social Work, Seinäjoki University of Applied Sciences Seinäjoki, Finland.
`pedro.morenosanchez@seamk.fi`
Phone: +358 403752546

Total words of manuscript: 4349 words
Total words of abstract: 207 words
Number of figures: 4 figures
Number of tables: 6 tables

# Development and evaluation of an Explainable Prediction Model for Chronic Kidney Disease Patients based on Ensemble Trees

Pedro A. Moreno-Sanchez [1[0000-0003-3987-1789]]

[1] School of Health Care and Social Work, Seinäjoki University of Applied Sciences Seinäjoki, Finland.
`pedro.morenosanchez@seamk.fi`

**Abstract.** Chronic Kidney Disease (CKD), where delayed recognition implies premature mortality, is currently experiencing a globally increasing incidence and high cost to health systems. Data mining allows discovering subtle patterns in CKD indicators to contribute to an early diagnosis. This work presents the development and evaluation of an explainable prediction model that would support clinicians in the early diagnosis of CKD patients. The model development is based on a data management pipeline that detects the best combination of ensemble trees algorithms and features selected concerning classification performance. Furthermore, the main contribution of the paper involves an explainability-driven approach that allows selecting the best predictive model maintaining a balance between accuracy and explainability. Therefore, the most balanced explainable predictive model implements an extreme gradient boosting classifier over 3 features (packed cell value, specific gravity, and hypertension), achieving an accuracy of 99.2% and 97.5% with cross-validation technique and with new unseen data respectively. In addition, an analysis of the model´s explainability shows that the packed cell value is the most relevant feature that influences the prediction results of the model, followed by specific gravity and hypertension. This small number of feature selected results in a reduced cost of the early diagnosis of CKD implying a promising solution for developing countries.

**Keywords:** medical XAI; clinical prediction model; Chronic Kidney Disease; feature selection; explainability.

## 1 Introduction

Chronic kidney disease (CKD) is manifested as a worldwide public health problem with increasing incidence and prevalence that leads an ample number of patients to premature mortality [1], and implies high cost to healthcare systems especially in developing countries where lack of appropriate treatment results in a high mortality rate within their population [2][3]. Typically, CKD is a disease with no early symptoms that when detected the kidney has already lost 25 per cent of its capacity and is under progressive damage that, if not slowed by controlling underlying risk factors (hypertension, obesity, heart disease, age, diabetes, drug abuse, family history of kidney disease, race/ethnicity) [4], the hemodialysis or even kidney transplantation are crucial for patient survival. [5–7]. Therefore, an early diagnosis of CKD based on those indicators allows initiating treatments to slow the progression of kidney damage and prolong patients' life even with an insufficient kidney function.

In the field of medicine, Artificial intelligence (AI) have become, in recent years, a promising instrument to build computer-aided diagnosis (CAD) [8], [9], that could be employed to discover latent correlations between CKD onset and its indicators enabling an early discover of those patients at risk. In the literature, several AI approaches for CKD detection employed their own data sources as medical images or tabular datasets of clinical indicators extracted from EHR [10]. Other researchers used a public common CKD dataset from the University of California Irvine-Machine Learning [11] to build their prediction models allowing reproducibility of results as well as benchmarking between other models implementation. For the sake of these latter aspects, Table 1 shows the most recent and accurate works (accuracy above 98%) that employed the CKD dataset of the UCI-ML repository [11] and applied feature selection approaches in their data preparation stages to build the prediction models.

When CAD systems' decisions affect patients' life eventually, explanations about the output's logic of the AI models are crucial to support clinicians in their diagnosis and treatments. Thus in the medical domain, eXplainable Artificial Intelligence (XAI) would allow healthcare experts to make reasonable and data-driven decisions as well as improve the clinical adoption of AI models [12]. Global model-specific solutions of XAI have been developed for the last decade in different clinical fields, namely: urology [13], toxicology [13], endocrinology [14], neurology [15], cardiology [16], cancer (e.g. breast cancer or prostate cancer) [17], [18], and chronic diseases (e.g. diabetes or Alzheimer's disease) [19, 20]. Concerning XAI application to prediction models, an inherent trade-off must be addressed between AI models' predictive accuracy and their explainability, because most accurate models are usually less transparent and vice versa.

This paper aims at describing the development and assessment of an explainable prediction model of CKD that address the balance between accuracy and explainability through an automated data pipeline that implements different ensemble trees algorithms and feature selection techniques to achieve the best accuracy. In addition, an explainability analysis is conducted in terms of feature relevance and explainability metrics. The outline of the article adopts the following structure: Section 2 gives an overview of the dataset as well as the machine learning algorithms, evaluation metrics and explainability techniques employed in this research. Section 3 presents the pipeline employed to build the prediction model, the evaluation results in terms of classification and explainability, and the explainability analysis. The discussion of the results, the comparison to the related works, as well as the potential impacts of the prediction model on the healthcare domain are described in Section 4. Finally, Section 5 includes the conclusions drawn from the work.

**Table 1.** Classification results (in %) of related works and machine learning classifiers (best ones in italic). (*Acc*: accuracy; *Sen*: sensivity; *Spe*: specificity; *F1*:f1-score; *Pre*: Precision; *#F*: number of features; *: Studies that perform the best classifier with unseen new data).

| Article | Acc | Sen | Spe | F1 | Pre | #F | Machine Learning Classifier |
|---|---|---|---|---|---|---|---|
| Ekanayake [21] | 100 | 100 | - | 100 | 100 | 7 | *DT, RF,* XGB, *Ada*, ET (*) |
| Alaoui [22] | 100 | - | - | - | - | 23 | *XGB Lin*, Lin SVM, DT, RF |
| Ogunleye [23] | 100 | 100 | 100 | - | 100 | 12 | *XGB* (*) |
| Zeynu [24] | 99.5 | 99.5 | - | 99.5 | 99.5 | 8 | KNN, DT, *ANN*, NB,SVM. |
| Raju [25] | 99.3 | 99 | - | 99 | 100 | 5 | XGB *RF*, LR, SVM,NB (*) |
| Khan [26] | 99.1 | 99.7 | - | 99.3 | 98.7 | 23 | *NB*, LR, SVM, DT, RF |
| Hasan [27] | 99 | - | - | 99 | - | 13 | *Ada*, RF, GB, ET(*) |
| Abdullah [28] | 98.8 | 98.0 | 100 | 98.8 | 98.0 | 10 | *RF*, SVM, NB, LR |
| Alaiad [29] | 98.5 | 99.6 | 96.8 | - | 98 | 12 | NB, DT, SVM, *KNN*, Jrip |
| Kadhum [30] | 98.1 | 98 | - | 98 | 98 | 10 | SVM, *ELM* |

## 2  Material and methods

### 2.1  Chronic Kidney Disease dataset

To promote the reproducibility of this research, the UCI-ML dataset was employed. Table 2 shows the structure of the dataset that includes 400 samples collected from the Apollo Hospitals, Karaikudi, India during a nearly 2-month period in 2015. Each instance of the dataset is composed of 11 numeric, 10 nominal, 3 ordinal features and 1 target feature (notckd/ckd).

### 2.2  Ensemble Trees machine learning techniques

Ensemble trees have become one of the most popular machine learning classifiers due to their stability and robustness when dealing with datasets of any size, as well as to a reasonably good predictive performance. Ensemble trees perform classification tasks by weighting various decision trees and combining them to reach a final model that improves each base model [31]. In addition, ensemble methods are used to mitigate challenges like class imbalance or the curse of dimensionality. Apart from decision trees, the classifiers used in this research are: random forest or extra trees [31] that follow the bagging technique where each base decision tree is trained using a sample with the same number of instances taken with replacement from the original dataset; and adaptative boosting [31] or extreme gradient boosting [32] that employ the technique of boosting focused on instances, in a sequential way, that have been previously misclassified when training a new base decision tree.

Table 2. Attributes description of ckd dataset

| Features (units) [*legend*] | Type of feature (classes) | Average (std) / number of values |
|---|---|---|
| Age (year) [*age*] | Numerical | 51.48 (17.17) |
| Blood pressure (mm/Hg) [*bp*] | Numerical | 76.46 (13.68) |
| Specific gravity [*sg*] | Ordinal (1.005,1.010,1.015, 1.020, 1.025) | 7, 84, 75, 106, 81 |
| Albumin [*al*] | Ordinal (0,1,2,3,4,5) | 199,44,43,43,24,1 |
| Sugar [*su*] | Ordinal (0,1,2,3,4,5) | 290,13,18,14,13,3 |
| Red blood cells [*rbc*] | Nominal (normal/abnormal) | 47 abnormal |
| Pus cell [*pc*] | Nominal (normal/abnormal) | 76 abnormal |
| Pus cell clumps [*pcc*] | Nominal (not present/ present) | 42 present |
| Bacteria [*ba*] | Nominal (not present/ present) | 22 present |
| Blood glucose random (mgs/dl) [*bgr*] | Numerical | 148.04 (79.28) |
| Blood urea (mgs/dl) [*bu*] | Numerical | 57.43 (50.50) |
| Serum creatinine (mgs/dl) [*sc*] | Numerical | 3.07 (5.74) |
| Sodium (mEq/l) [*sod*] | Numerical | 137.53 (10.41) |
| Potassium (mEq/l) [*pot*] | Numerical | 4.63 (3.19) |
| Hemoglobin (gms) [*hemo*] | Numerical | 12.53 (2.91) |
| Packed cell volume [*pcv*] | Numerical | 38.88 (8.99) |
| White blood cell count (cells/cumm) [*wc*] | Numerical | 8406.12 (2944.47) |
| Red blood cell count (cells/cumm) [*rc*] | Numerical | 4.71 (1.03) |
| Hypertension [*htn*] | Nominal (no/yes) | 147 yes |
| Diabetes mellitus [*dm*] | Nominal (no/yes) | 137 yes |
| Coronary artery disease [*cad*] | Nominal (no/yes) | 34 yes |
| Appetite [*appet*] | Nominal (good/poor) | 82 poor |
| Pedal edema [*pe*] | Nominal (no/yes) | 76 yes |
| Anemia [*ane*] | Nominal (no/yes) | 60 yes |
| Target class | notckd/ckd | 250 ckd |

## 2.3 Explainability techniques for ML

Concerning explainability, in domains (e.g healthcare) where predictions results must be interpretable, classifiers like decision trees are preferred due to their transparency. However, albeit ensemble trees usually present a better performance in classification tasks than transparent models, they require post-hoc explainability techniques to interpret their results because of their black-box behaviour. Post-hoc explainability techniques aim at providing understandable information about how an already developed model produces its predictions [33]. In this work, the following techniques have been used: *implicit feature importance*, which measures the mean decrease impurity, known as Gini, for each feature when obtaining the prediction; *feature permutation importance*, which quantifies the prediction error increase of the model after permuting a specific feature's values, being the most important features those that provoke an error increase [34]; *partial dependence plot*, that shows visually the marginal effect in terms of the probability that a given feature has on the predicted outcome over a range of different observed values, while all other features are kept constant [35]; and *SHapley Additive exPlanations (SHAP)* that computes by applying coalitional game theory, an additive importance score for each feature in every individual prediction with local accuracy and consistency which are aggregated to give a global explainability of the model [36] [37].

Besides these explainability techniques, it's worth mentioning that feature selection procedures can be applied to remove unimportant features that bring non-relevant information to the classification, hemce enhancing models' explainability [38]. This research has embraced feature selection by applying filter methods, where intrinsic properties of data, measured with ANOVA, Chi-squared or mutual information test, justify the inclusion of an attribute or a subset of attributes; or through wrappers methods like Recursive Feature Elimination (RFE) where a classification algorithm is utilized to select important features.

### 2.4 Classification performance and explainability metrics

Since the dataset employed presents an imbalance in its target feature (250 CKD/150 non-CKD) other metrics than accuracy, that measure the rate of true predictions over the total, are needed, namely: sensitivity (fraction of positive instances predicted correctly); specificity (fraction of negative instances predicted correctly); precision (fraction of true positive data given all true predicted data), and F1-score (harmonic mean from precision and sensitivity) [13]. The formulas of these metrics are shown in Table 3.

Moreover, considering ensemble trees as the classifiers employed, the explainability metrics proposed by Tagaris et.al [39] are used: *Interpretability*, defined as the percentage of those masked features that do not bring information to the final classification result and the total number of features of the dataset; Fidelity, that measures the accuracy relation of the equivalent full-interpretable model (i.e. decision trees) and its un-interpretable model counterpart; and Fidelity-Interpretability Ratio (FIR) that shows how much of the model's interpretability is sacrificed for performance, being 0.5 the optimal score.

Table 3. Metrics' formula (TN: true negative, FN: false negative, FP: false positive, TP: true positive).

| Metrics | Formula | |
|---|---|---|
| *Accuracy (Acc)* | (TP+TN) / (TP+TN+FP+FN) | (1) |
| *Sensitivity/Recall (Sen)* | TP / (TP+FN) | (2) |
| *Specificity (Spe)* | TN / (FP+TN) | (3) |
| *Precision (Pre)* | TP / (TP+FP) | (4) |
| *F1-Score (F1)* | 2*((Precision*Recall) / (Precision+Recall)) | (5) |
| *Interpretability (I)* | I= masked features/total input features | (6) |
| *Fidelity (F)* | F=Acc. $_{\text{Fully Interpretable}}$ / Acc. $_{\text{Un-Interpretable}}$ | (7) |
| *Fidelity-Interpretability Ratio (FIR)* | FIR=F/(F+I) | (8) |

## 3 Results

### 3.1 Automated data workflow pipeline

In this work, the SCI-XAI automated data workflow pipeline published in [40] has been employed for developing the explainable CKD prediction model. Figure 1 shows the SCI-XAI pipeline that is based on the python scikit-learn package [41] and allows through a brute force algorithm finding the specific combination of ensemble trees classifier, the number of features selected, the feature selection and data missing imputation methods that output the best classification performance of the model. As a first step, the original dataset is split, with a target feature stratification, into training and test set with 280 and 120 instances respectively (ratio 70/30). This initial split allows building the model exclusively with training set's instances, and hence, evaluating its performance over unseen new data from the test set, when the abovementioned preprocessing and classification discovered parameters are applied. Next, the data preparation phase takes place separately for numerical, nominal, and ordinal features to be merged in the modelling or training phase, where through a 5-fold cross-validation approach different ensemble trees classifiers are applied.

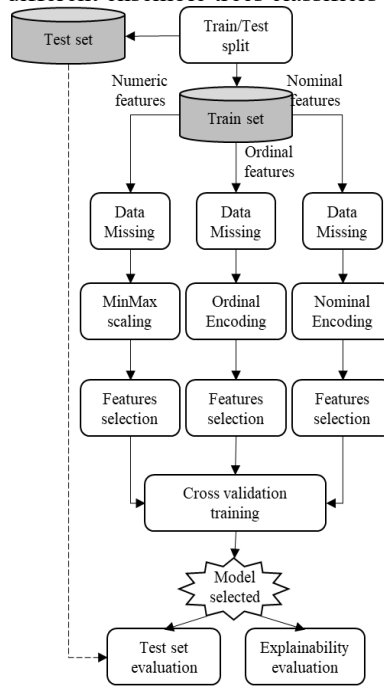

Figure 1. SCI-XAI automated data workflow pipeline.

### 3.2 Feature selection

The best combination of features selected found by SCI-XAI pipeline for each ensemble trees algorithms is shown in Table 5. The number of features, the techniques employed for selection (ANOVA, Chi-squared, mutual information or Recursive Feature Elimination), and the name of those selected features are detailed for numerical, nominal and ordinal features of the dataset. Thus, the feature selection step denotes at least 50% of the original features a non-relevant for the classification output. XGBoost is the algorithm with the biggest feature reduction in its best classification results, remaining with 3 out of 24 features.

### 3.3 Classification and explainability metrics results

Table shows the classification performance of the different ensemble trees algorithms after the training cross-validation module as well as the evaluation with the test set. The results show a solid classification performance in the training phase with a range of 98.1 to 100% in all metrics considered. This robust performance is maintained with unseen data from the test, obtaining in all classifiers considered an accuracy of more than 97.5%, and above 95% in the rest of the classification metrics.

Once the number of relevant features selected is known, the evaluation of explainability can be performed (shown in Table ). For each ensemble trees algorithm and its group of selected features, a specific decision tree was implemented to obtain the equivalent fully interpretable model to calculate the Fidelity. As FIR gives a balanced measure between interpretability and fidelity with 0.5 as the optimal point, XGBoost (FIR=0.53) achieved the most balanced model. Therefore, the XGBoost and its group of selected features are used to conduct an explainability analysis of its predictions.

### 3.4 Explainability analysis of the prediction model

Since the CKD predicition model based on XGBoost is selected as the most balanced model in terms of explainability and accuracy, in this subsection, the relevance of the following features pcv (packed cell volume), htn (hypertension), sg (specific gravity) is analysed with different post-hoc explainability techniques to show their influence in model's outputs.

The implicit feature importance and feature permutation importance techniques allow visualizing the global explainability of each model's feature without informing about the direction of the contribution, i.e. increase or decrease of CKD probability. Figure 2.a shows the ranking of the features selected's importance based on mean impurity decrease (Gini) by applying feature implicit importance. In addition, Figure 2.b shows those features' importance, using the feature permutation importance technique, calculated as the decrease in accuracy when a specific feature' values are permuted. Both techniques denote pcv as the most relevant feature followed by sg and htn (in descending order of importance).

When employing PDP, information about the contribution that different features' values have to the probability when predicting a positive case of CKD can be extract-ed. Figure 3 shows the PDP plots of the three selected features where the x-axis represents the values distribution, and the y-axis the contribution in probability to the prediction of CKD. Thus, for pcv values above 39 the contribution to predicting CKD decrease from 1 to 0.6, being monotonic at 0.6 for the rest of values except at 41 where a small peak can be appreciated. In addition, patients with hypertension (htn=1) have an increase of 0.4 (from 0.6 to 1.0) in the probability of suffering CKD. In the case of sg, for values of 1.020 and 1.025, the feature contributes to reducing the probability of predicting CKD in a 0.4 (from 1.0 to 0.6). It could be appreciated that approximately 0.6 represents a base value in the partial dependence plots which correspond with the proportion of total CKD positive (250) and negative (150) cases in the dataset.

Table 4: Feature selection results (#: number of features selected; Feats: name of features selected; mut-inf: mutual information, RFE: Recursive Feature Elimination)

| Classifier | Numerical features | | Nominal features | | Ordinal features | | Total of feats. |
|---|---|---|---|---|---|---|---|
| | # | Feats (selection method) | # | Feats (selection method) | # | Feats (selection method) | |
| *Decision Trees* | 1 | hemo (ANOVA) | 3 | htn, dm, appet (chi2) | 1 | sg (mut-inf) | 5 |
| *Random Forest* | 1 | hemo (ANOVA) | 5 | (RFE) | 1 | sg (mut-inf) | 7 |
| *Extra Trees* | 4 | pcv, hemo, rc, sc (mut-inf) | 3 | htn, dm, appet (chi2) | 1 | sg (mut-inf) | 8 |
| *AdaBoost* | 7 | pcv, hemo, rc, sc, pot, sod, bu (mut-inf) | 4 | htn, dm, appet, pe (mut-inf) | 1 | sg (mut-inf) | 12 |
| *XGBoost* | 1 | pcv (mut-inf) | 1 | htn (mut-inf) | 1 | sg (mut-inf) | 3 |

Table 5. Classification metrics results (in %)

| Classifier | Training set (Cross-val approach) | | | | | Test set (new unseen data) | | | | |
|---|---|---|---|---|---|---|---|---|---|---|
| | Acc. | Sens. | Spec. | F1 | Prec. | Acc. | Sens. | Spec. | F1 | Prec. |
| *Decision Trees* | 99.6 | 99.4 | 100 | 99.7 | 100 | 97.5 | 96 | 100 | 98 | 100 |
| *Random Forest* | 100 | 100 | 100 | 100 | 100 | 97.5 | 96 | 100 | 98 | 100 |
| *Extra Trees* | 100 | 100 | 100 | 100 | 100 | 98.3 | 97.3 | 100.0 | 98.6 | 100.0 |
| *AdaBoost* | 100 | 100 | 100 | 100 | 100 | 98.3 | 97.3 | 100.0 | 98.6 | 100.0 |
| *XGBoost* | 99.2 | 100 | 98.1 | 99.4 | 98.8 | 97.5 | 98.7 | 95.6 | 98 | 97.4 |

Table 6. Explainability metrics results

| Classifier | Interpretability | Fidelity | FIR |
|---|---|---|---|
| *Decision Trees* | 79 % | 100 % | 0.56 |
| *Random Forest* | 71 % | 100 % | 0.59 |
| *Extra Trees* | 67 % | 99 % | 0.60 |
| *AdaBoost* | 50 % | 99 % | 0.66 |
| *XGBoost* | 88 % | 100 % | 0.53 |

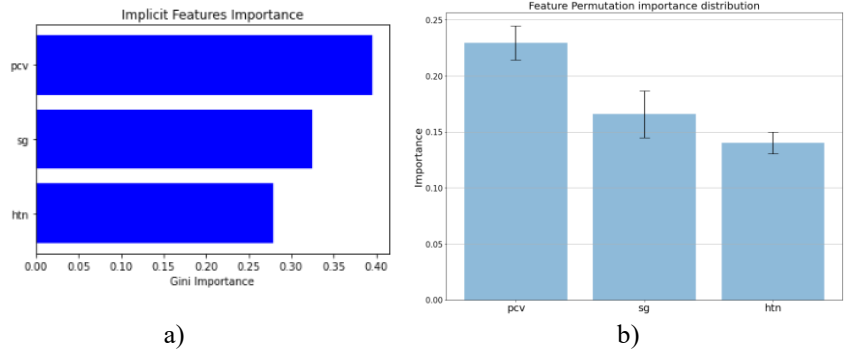

Figure 2: Global explainability of prediction model (a. Implicit feature importance, b. Feature permutation importance)

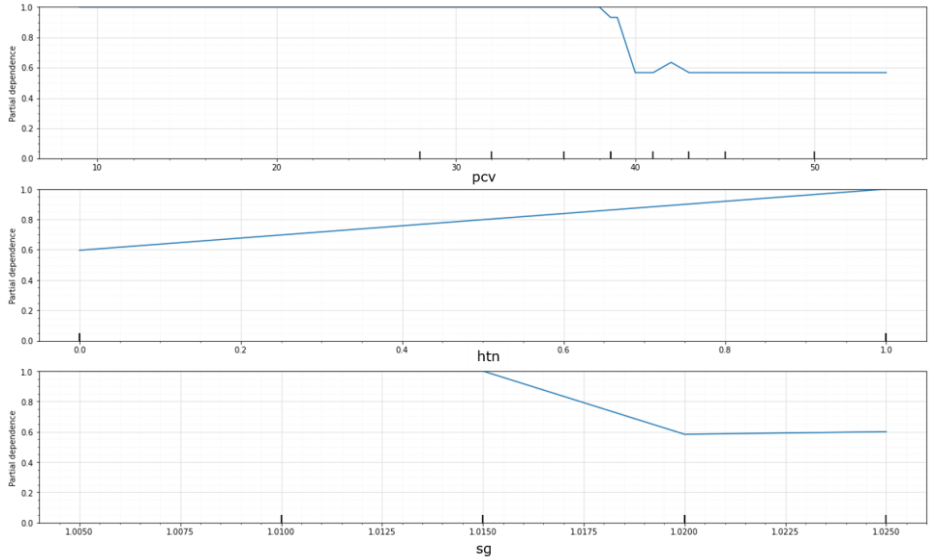

Figure 3: PDP plots for CKD probability contribution of each model's feature.

As an example of post-hoc explainability techniques when concerning predictions of individual cases, the SHAP technique depicts through its "waterfall plots", the attribution of each feature value not only specifying the direction force towards the final score (red: positive contribution, blue: negative contribution) but also the feature's weight (length of the bar). Thus, Figure 4.a and Figure 4.b show the features' attributions when predicting respectively a true negative case (y=0, the patient does not have CKD) and a true positive case (y=1, the patient does have CKD). For both cases, the prediction starts from a base SHAP value (1.66) equal to the initial distribution of CKD/nonCKD cases (250/150). In the case of the true negative, pcv, with a value of 47, reveals as the most relevant feature in the prediction with an attribution of -3.09, meanwhile sg and hth, with values 1.025 and 0 respectively, has negative attributions towards the non-CKD prediction (-1.92 and -1.44). As regards the true positive case, the values of pcv=35, sg=1.015 and htn=1 contribute to a positive prediction of CKD with a nearly similar additive attribution (+1.43, +1.38, +1.22 respectively). It is worth noting that the contributions shown for the features' values in these individual cases agree with the insights gained with the PDP plots.

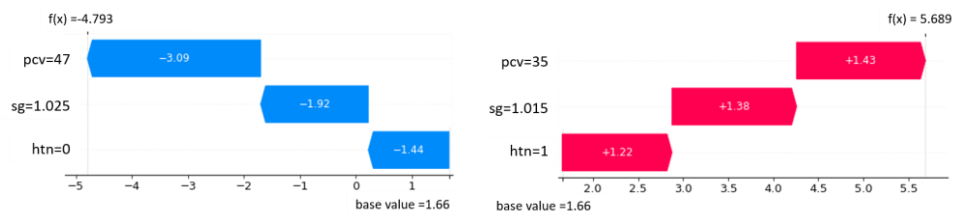

Figure 4. Local Explainability through SHAP (a. True Negative case; b. True Positive case)

## 4      Discussion

Due to the current increase of global incidence of CKD, the classification of those patients who might suffer the disease using prediction models could become a relevant tool for doctors to achieve an early diagnosis. In addition to that, XAI could imply an improvement to those prediction models by meeting the healthcare professionals' demands about understanding the decisions made by the models. Having more explainable CKD prediction models, doctors could make more data-driven decisions and focus on controlling those underlying features or indicators to slow the progressive damage of the kidney

This paper describes a CKD prediction model developed to tackle the early diagnosis not only seeking high accuracy of the predictions but also analysing the explainability of its results. Thus, this research contributes to enlarging the works dedicated to CKD diagnosis through AI from a novelty perspective, to the best of our knowledge, that focus on the model's explainability. By using post-hoc explainability techniques, this work aims at "opening" the black-box paradigm of the ensemble trees classifiers employed to build the CKD prediction model.

The development of the explainable CKD prediction model is based on a data management pipeline, previously developed in other author's work, that automates the data preparation, modelling and evaluation phases. The pipeline allows inferring automatically different parameters like the appropriate ensemble tree algorithm, the relevant features selected, the feature selection method, and data imputation techniques to obtain the best classification performance of the prediction model. Moreover, the pipeline allows evaluating the model's performance over new unseen data (30% of the original dataset), which could emulate a real deployment, however, the model's performance in a real clinical environment might differ from the results because the medical records are not usually as curated as the dataset employed.

Concerning the classification performance, different ensemble tree algorithms have been considered (i.e. random forest, extra trees, AdaBoost, and XGBoost) along with different combinations of features selected to build the prediction model and consequently to conduct its evaluation. Considering the classification results achieved, this work achieves the state-of-art of CKD prediction models found in the literature. Thus, the application of the SCI-XAI pipeline shows a fairly good performance, especially when compared to other related works.

The pipeline´s feature selection step has also proven to be valuable due to the substantial reduction of the original number of features, leaving 3 out of 24 when using the XGBoost classifier, which is the best CKD prediction model in the literature in terms of minimum features considered. Furthermore, 4 out 5 ensemble learning algorithms considered obtaining their best classification results with only 33% of the original features showing the capability of the pipeline to detect relevant features when building the prediction model.

To the best of our knowledge, this paper is the first to entail an explainability analysis of a CKD prediction model. By considering the FIR value of 0.5 as the optimal point in terms of a trade-off between accuracy and explainability, the model that uses the XGBoost is selected as the most balanced. By doing so, this selected model does not achieve the best classification performance, showing an example of the tension occurring between accuracy and explainability when concerning prediction models aimed at being used in specific domains where understanding the results is crucial.

Regarding the analysis of the features' importance in the prediction model, the pcv (packed cell volume) feature seems to be the most relevant in all post-hoc analysis techniques considered, followed by the sg (specific gravity) and then htn (hypertension). It is worth highlighting the utility of the PDP plots to identify thresholds on which a certain feature modifies the probability prediction. For instance, this work establishes thresholds in 37 and 1.015 for pcv and sg respectively from the probability starts to decrease, implying that doctors could set up a treatment for the patient to be above these thresholds and reduce the probability of CKD disease. Moreover, the local explainability results exemplify how explainability techniques contribute to the promotion of precision or personalized medicine by showing the relevance of the different features for an individual prediction case.

With the results described in this work, the added value of explainability to a clinical prediction model is exhibited. Moreover, the feature selection approach is valuable not only for improving the explainability of clinical prediction models but also for reducing the cost of the diagnosis having fewer clinical indicators to extract. In particular, since this explainable CKD prediction model embraces the processing of 3 features (pcv, sg, and htn), the cost associated to extract them, by following the price list defined by Salekin et al [42], is 1.62 USD for pcv in a hematocrit test, free for sg, and free for htn. Therefore, the cost associated with an early diagnosis of CKD by using this explainable prediction model would be around 1.6 USD, which would have an important impact on developing countries where medical access is more difficult [43].

## 5     Conclusions

The development and evaluation of an explainable CKD prediction model have been presented in this work with the aim of showing the importance of considering XAI in early diagnosis systems in the healthcare field, maintaining a balance between the classification performance of the model and its explainability. The prediction model is built by using a data management pipeline that allows the detection of the most adequate ensemble tree algorithm as well as the number of features selected. In order to detect the best-balanced model in terms of accuracy and explainability, different evaluations are carried out by applying classification and explainability metrics. Therefore, the best explainable prediction model implements an XGBoost classifier over the following 3 features: packed cell values (pcv), specific gravity (sg) and hypertension (htn). After an explainability analysis by employing different techniques, the features' relevance in descendent order is: pcv, sg and htn. The prediction model developed equals the classification performance of the best CKD prediction models identified in the literature. In addition, the novelty presented by this work is the explainability approach adopted in the model's performance which revolves around giving healthcare professionals an easier understanding and interpretability of the outcomes generated by the model. Thus, not only would clinicians achieve an early diagnosis with a reduced group of indicators, but they could also focus on tackling relevant features to avoid the CKD onset or even to revert its progress. Future works would also be aimed at testing the prediction model developed in a clinical setting to test the accuracy robustness of the model with new patients' data as well as to gather insights of healthcare professionals about the explainability of the results provided.

**Glossary of Terms**

| | |
|---|---|
| • DT: Decision Trees<br>• RF: Random Forest<br>• XGB: eXtreme Gradient Boosting<br>• Ada: Adaptive Boosting<br>• ET: Extra Trees<br>• XGB lin: XGB linear<br>• Lin SVM: Linear Support Vector Machine<br>• KNN: K-Nearest Neighbors<br>• ANN: Artificial Neural Network<br>• NB: Gaussian Naïve Bayes<br>• LR: Logistic Regression<br>• GB: Gradient Boosting<br>• Jrip: Jrip associated rule<br>• ELM: Extreme Marchine Learning<br>• Acc: Accuracy<br>• Sen: Sensivity | • Spe: Specificity<br>• F1: f1-score<br>• Pre: Precision<br>pcv: Packed cell volume<br>hemo: Hemoglobin (gms)<br>rc: Red blood cell count (cells/ cumm)<br>sc: Serum creatinine (mgs/dl)<br>pot: Potassium (mEq/l)<br>sod: Sodium (mEq/l)<br>bu: Blood urea (mgs/dl)<br>htn: Hypertension<br>dm: Diabetes mellitus<br>appet: Appetite<br>rbc: Red blood cells<br>pc: Pus cell<br>pe: Pedal edema<br>sg: Specific gravity |

**Author's Biography**

| 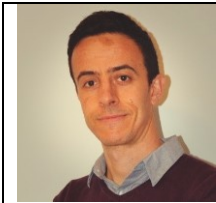 | **Pedro A. Moreno-Sanchez** currently works as Researcher at Seinäjoki University of Applied Sciences, Finland. He received the Ph. D. on 2014 by the Polytechnical University of Madrid, Spain. His main research interest are eXplainable Artificial Intelligence applied to healthcare. |
|---|---|